\documentclass[10pt,twocolumn,letterpaper]{article}

\usepackage{iccv}
\usepackage{times}
\usepackage{epsfig}
\usepackage{graphicx,subfigure}
\usepackage{amsmath}
\usepackage{amssymb}

\usepackage{comment}

\usepackage[pagebackref=true,breaklinks=true,letterpaper=true,colorlinks,bookmarks=false]{hyperref}

\iccvfinalcopy 

\def\iccvPaperID{2934} 

\ificcvfinal\fi
\begin{document}

\title{Harvesting Information from Captions for Weakly Supervised Semantic Segmentation}

\author{Johann Sawatzky
	\qquad
	Debayan Banerjee
	\qquad
	Juergen Gall\\
	University of Bonn \\ {\tt\small \{jsawatzk, debayan, jgall\} @ uni-bonn.de}
}

\maketitle

\begin{abstract}
Since acquiring pixel-wise annotations for training convolutional neural networks for semantic image segmentation is time-consuming, weakly supervised approaches that only require class tags have been proposed. In this work, we propose another form of supervision, namely image captions as they can be found on the Internet.
These captions have two advantages. They do not require additional curation as it is the case for the clean class tags used by current weakly supervised approaches and they provide textual context for the classes present in an image.
To leverage such textual context, we deploy a multi-modal network that learns a joint embedding of the visual representation of the image and the textual representation of the caption. The network estimates text activation maps (TAMs) for class names as well as compound concepts, \ie combinations of nouns and their attributes. 
The TAMs of compound concepts describing classes of interest substantially imrpove the quality of the estimated class activation maps which are then used to train a network for semantic segmentation. 
We evaluate our method on the COCO dataset where it achieves state of the art results for weakly supervised image segmentation. 

\end{abstract}

\section{Introduction}





\begin{figure}[t]
 \includegraphics[width=80mm]{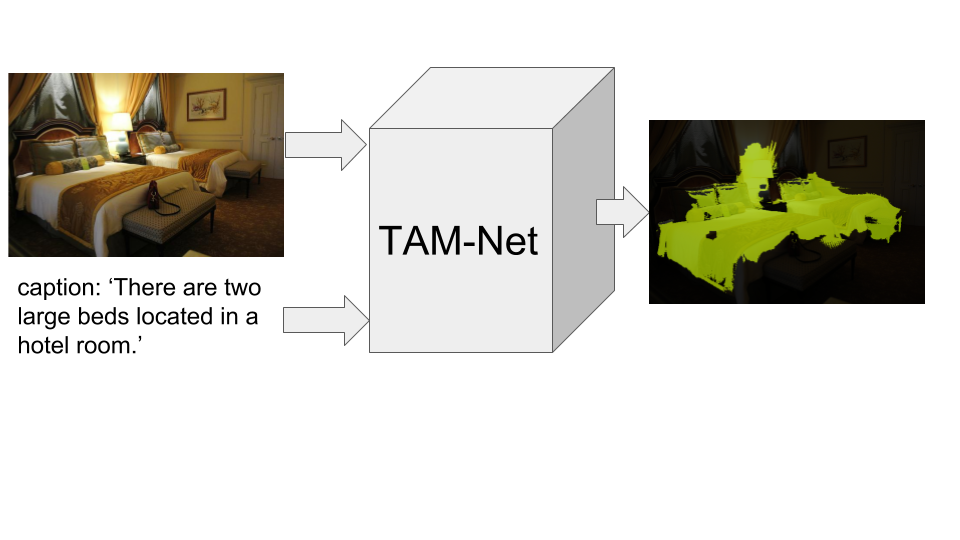}
 \caption{Given one or multiple captions per image in the training set, our network predicts text activation maps (TAMs) for each image which are then converted into class activation maps (CAMs) as illustrated in Figure~\ref{fig:inf}. 
 The text activation maps are more general than class activation maps since they localize compound concepts like `two large beds' as well as categories like `bed'. 
 The example shows the class activation map estimated for `bed' for this training example. Using the estimated CAMs of all training images, a standard convolutional neural network for semantic segmentation can then be trained.}
 \label{fig:intro}
 \vspace{-5mm}
\end{figure}

Fully convolutional networks have shown to perform very well on the task of semantic image segmentation, but training such networks requires data annotated on a pixel level. Obtaining such annotations, however, is very time consuming and expensive. 
For this reason, approaches for weakly supervised semantic segmentation have been proposed that require less supervision for training. While the type of supervision ranges from class tags, key points, scribbles to bounding boxes, the vast majority of these approaches rely on class tags since it is assumed that these tags can be more easily acquired. A popular approach for weakly supervised learning from image tags~\cite{SEC,CAM,affinityNet} consists of estimating for each image in the training set so-called class activation maps (CAMs) \cite{CAM}, which indicate where certain semantic categories occur in an image. In a second step, pixelwise class labels are extracted from the CAMs and a neural network for image segmentation is trained. These approaches, however, still assume curated image tags, \ie all classes of interest in the image must be tagged. This is not guaranteed for tags retrieved together with images from the Internet. 

Another form of supervision are readily available textual descriptions. Such textual descriptions can be either from image captions or text surrounding or referring to an image in an article or blog. Such textual descriptions are more general than class tags and it is possible to reduce these textual descriptions to class tags by parsing the texts for the names of the categories of interest. However, 
textual descriptions are richer in the description of the image content than just class tags as illustrated in Figure~\ref{fig:intro}. In case of class tags, the image would be only labeled by the relevant category `bed' and the only information we have is that the image contains at least one bed, but we do not know if the bed is large or small or if many instances are in the image. The caption ``There are two large beds located in a hotel room'' provides much more information. In particular, `two large beds' indicates that many pixels of the image should be labeled by the category `bed'.    

In this work, we therefore propose an approach that uses non curated image captions as weak supervision for training a convolutional neural network for semantic image segmentation. Our contribution focuses on the research question how class activation maps (CAMs) can be estimated from image captions and we show that these class activation maps are substantially more accurate than CAMs estimated from class tags. 
In order to estimate the class activation maps for the training images from captions, we learn a joint embedding of the visual representation of the image and the textual representation of the caption as illustrated in Figure~\ref{fig:inf}. 
Using such a joint embedding, our network predicts text activation maps (TAMs), which locate categories like `bed' as well compound concepts like `two large beds'. For each class, the TAM of the class as well as the TAMs of the compounds which contain the class name are fused to generate the CAM for the class.  

We provide a thorough ablation study that analyses the benefit of the additional textual context that is provided by the compound concepts. On the COCO dataset \cite{COCO}, the proposed approach outperforms the current state of the art in weakly supervised semantic segmentation.

\section{Related Work}
Fully supervised semantic segmentation has been studied in many works,
e.g., \cite{deeplab}, \cite{paper27}, \cite{paper29}, \cite{paper46}, \cite{paper47}.
More recently, weakly-supervised semantic segmentation has
come to the fore. Early work such as \cite{paper41} relied on hand-crafted features, such as color,
texture, and histogram information to build a graphical model. However, with the advent
of convolutional neural networks (CNN), this conventional approach has been
gradually replaced because of its lower performance on challenging benchmarks \cite{paper11}.

A natural step to less supervision are more coarse spatial cues like bounding boxes, key points and scribbles.
In \cite{paper32}, Papandreou et al.use the expectation-maximization algorithm to perform
weakly-supervised semantic segmentation based on annotated bounding boxes
and image-level labels. Another more sophisticated approach based on bounding boxes was proposed in \cite{khoreva_CVPR17}. The authors use region proposal generated by Multiscale
Combinatorial Grouping (MCG) \cite{paper35} and Grabcut \cite{Grabcut} to localize the objects more precisely within the bounding box.
More recently, Li et al.~\cite{Li_2018_ECCV} used bounding boxes for object classes and image level supervision for stuff classes.
In \cite{paper26}, Lin et al.made use of a region-based graphical model, with scribbles providing ground-truth
annotations to train the segmentation network. Scribbles also served as supervision for the works of \cite{Tang2018Normalized, Tang_2018_ECCV}, which investigate loss regularizations.
Human annotated keypoints were used by Bearman et al.~\cite{paper2} for weakly supervised class segmentation and by Sawatzky et al.~\cite{Sawatzky_2017_CVPR} for weakly supervised affordance segmentation.

While the works mentioned above require some type of explicit spatial hints, others only rely on the list of present classes in the image.   
Qi et al.~\cite{paper36} used proposals generated by (MCG) \cite{paper35} to localize semantically meaningful
objects. Recently, Fan et al.~\cite{Fan_2018_ECCV} leveraged saliency to obtain object proposals and link objects of the same class across images with a graph partitioning approach. 
Pathak et al.~\cite{paper33} addressed the
weakly-supervised semantic segmentation problem by introducing a series of constraints.

In absence of explicit location cues provided by humans, class activation maps (CAMs) \cite{CAM} proved to be a seminal supervision source.
Pinheiro et al.~\cite{Pinheiro2015From, Shimoda2016Distinct} pioneered in this area. In \cite{SEC},
three loss functions are designed to gradually expand the high confidence areas of CAMs. This approach was first improved by Wei et al.~\cite{paper42} who use an adversarial erasing scheme
to acquire more meaningful regions that provide more accurate heuristic cues for training.
Recently, Huang et al.~\cite{paper34} proposed deep seeded region growing of CAMs with image level supervision.
In \cite{paper43}, Wei et al. presented a simple-to-complex framework which used saliency
maps produced by the methods \cite{paper6} and \cite{paper21} as initial guides. Hou et al.~\cite{paper19} advanced
this approach by combining the saliency maps \cite{paper18} with attention maps \cite{paper45}. 
Oh et al.~\cite{paper31} and Chaudhry et al.~\cite{paper4} considered linking saliency and attention
cues together, but they adopted different strategies to acquire semantic objects. Roy and
Todorovic \cite{paper38} leveraged both bottom-up and top-down attention cues and fused them
via a conditional random field as a recurrent network.
Ahn et al.~\cite{affinityNet} use image level class labels to generate an initial set of CAMs and then propagate those CAMs by using random walk predictions from AffinityNet.  
Wei et al.~\cite{Wei_2018_CVPR} use image level supervision with dilated convolutions with varying levels of dilations to generate weakly supervised segmentations. 
Wang et al.~\cite{Wang_2018_CVPR} use image level supervision along with a bottom-up and top-down framework, which alternatively expands object regions and optimizes segmentation network. 
In Briq et al.~\cite{Briq2018ConvolutionalSP} a convolutional simplex projection network is used for weakly supervised image segmentation. Tang et al.~\cite{Tang_2018_ECCV} integrate standard regularizers directly into the loss
functions over partial input for semantic segmentation.
Ge et al.~\cite{Ge_2018_CVPR} use a four stage process that combines object localization with filtering and fusion of object categories.
Hong et al.~\cite{paper17} and Jin et al.~\cite{paper22} tackle
the weakly-supervised semantic segmentation problem using images or videos from the Internet.
While the mentioned works mainly focus on steps following the generation of class activation maps, we focus on the CAMs themselves by leveraging object attributes from the captions.
Note that while the class tags are typically expected to be clean, \ie require human curation, our captions are not expected to mention all relevant classes.
This is closer to the realistic scenario of retrieving supervision information from the internet.

The task of weakly supervised visual grounding~\cite{Chen_2018_CVPR,Bouritsas_2018_CVPR,Zhao_2018_CVPR,DBLP:journals/corr/abs-1804-01720,Durand_2017,Xiao_2017} is related but a different task. Instead of training a network for semantic image segmentation, the goal is to localize a given phrase in an image and the challenge is to handle phrases that are not part of the training data. This means that they require a phrase for inference, while in our case the captions are only given for training but not for inference on the test dataset.


\section{Method}
Generating class activation maps \cite{CAM} on training images is a crucial intermediate step for the majority of current state of the art weakly supervised image segmentation methods. In our work, we therefore focus on improving them. 
To this end, we harvest information from image captions instead of relying on class tags only. Our TAM-network is a multimodal architecture, which maps image pixels and text snippets into a common semantic space which allows us to calculate activation maps for class names as well as compound concepts that include the class name as illustrated in Figure~\ref{fig:inf}. From TAMs of class names and class relevant compound concepts, we obtain class activation maps.
From these CAMs, we directly estimate the pixelwise class labels as described in Section~\ref{sec:inf}.
We finally train the widely used VGG16-deeplab model \cite{deeplab} for semantic segmentation on the estimated labels which yields us the final segmentation model.


\subsection{Parsing Captions}
\begin{figure}
 \includegraphics[width=80mm]{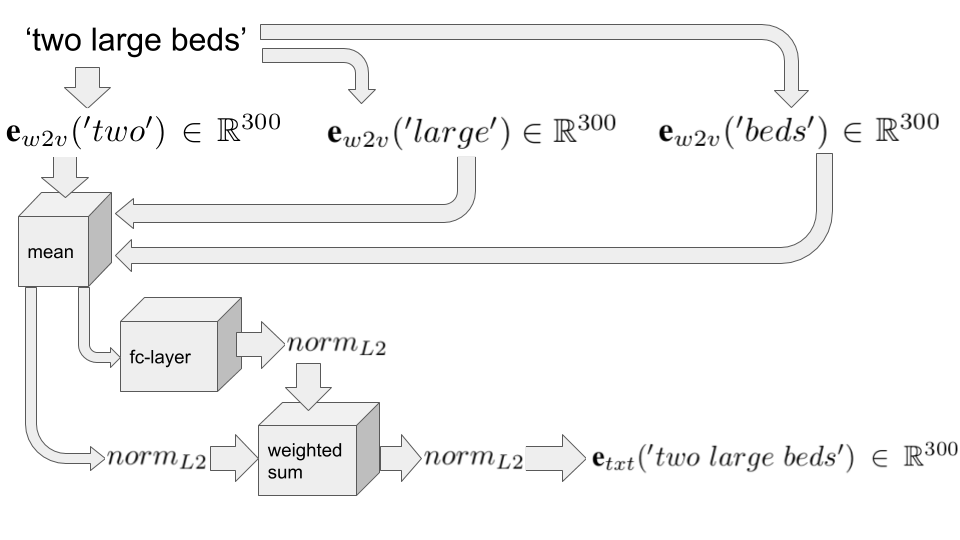}
 \caption{To obtain the textual embedding of an arbitrary snippet of text, we first encode each word with a word2vec model and average over words, which gives us the input embedding.
 Feeding it into a single fully connected layer with a residual connection yields the output embedding}
 \label{fig:txtemb}
 \vspace{-5mm}
\end{figure}
In our work, we distinguish three different types of text snippets: \textit{Class names} contain the names of the classes of interest in the particular dataset as well as their plural form.
\textit{Compound concepts} are snippets of a sentence between beginning of sentence, prepositions, verbs and end of sentence. These snippets have to contain multiple words excluding articles. Essentially these are combinations of numbers, adjectives, adverbs and nouns like \textit{two completely black dogs}. 
They are split into two categories: \textit{Class related compound concepts} contain a class name inside them and \textit{class unrelated compound concepts} do not.
All of them are used during training of the TAM network, but the type of the snippet determines its weight in the loss. For CAM inference, we use class names and class related compound concepts.
We use 300-dimensional word~to~vec (w2v) \cite{w2v} embeddings to convert text snippets to numerical vectors of equal length. The embedding of a single word is given by the word~to~vec dictionary. For text snippets containing multiple words, we take the arithmetical mean of the normalized embeddings of individual words.
These embeddings are used as input to the textual path of the TAM-network. 

We use the classes mentioned in the captions to determine what classes are present in a training image. If the class name is present in at least one of the image captions, the class is considered as present, otherwise not. Note that in contrast to curated image tags, the captions do not necessarily contain all classes that are present in an image.  

\subsection{Multi-Modal TAM Network}
Our TAM network comprises a visual path and a textual path which map visual and textual information into a common 300-dimensional semantic embedding space.

\textbf{Visual Path.}
Our visual embedding path maps an image $I$ with $P$ pixels to a pixelwise visual embedding $\textbf{E}_{vis}(I)\in\mathbb{R}^{P\times300}$.
It is a modification of the VGG16 architectures, but we will also report results for a ResNet38 architecture.
In both cases, we change the output dimension of the last fully connected layer to the dimension of the common semantic embedding space. 

\textbf{Textual path.}
As shown in Figure~\ref{fig:txtemb}, our textual path first obtains the word~to~vec embedding $\textbf{e}_{w2v}(t)\in\mathbb{R}^{300}$ of a text snippet $t$ by taking the average of the word~to~vec embeddings of the single words. 
Then $\textbf{e}_{w2v}(t)\in\mathbb{R}^{300}$ is mapped to a vector $\textbf{e}_{txt}(t)\in\mathbb{R}^{300}$ in the common semantic embedding space via:
\begin{equation}\label{eq:txt}
\begin{split}
 \textbf{e}_{txt}(t) = & norm_{L2}(norm_{L2}(\textbf{e}_{w2v}(t)) \\
                        &+ w_{res}norm_{L2}(\textbf{M}_{txt}\textbf{e}_{w2v}(t)))
\end{split}
\end{equation}
where $norm_{L2}()$ denotes normalization by the $L_2$ norm, $\textbf{M}_{txt}\in\mathbb{R}^{300\times300}$ is the weight matrix of a fully connected layer and $w_{res}\in\mathbb{R}$ is a hyperparameter. 
%
%
We also investigated RNNs that take the word~to~vec embeddings of the individual words as input and output $\textbf{e}_{txt}$, but they performed slightly worse than our approach. This is probably due to short length of our text snippets which contain mostly 2 or 3 words.

\textbf{Textual Activation Map.}
Given an image $I$ with $P$ pixels and a text snippet $t$, we generate the textual activation map (TAM), which we denote by $\textbf{x}(I,t)\in\mathbb{R}^{P}$, from the visual embedding $\textbf{E}_{vis}(I)\in\mathbb{R}^{P\times300}$ and the textual embedding $\textbf{e}_{txt}(t)\in\mathbb{R}^{300}$ by:
\begin{equation}\label{eq:TAM}
 \textbf{x}(I,t) = \textbf{E}_{vis}(I)\textbf{e}_{txt}(t). 
\end{equation}
To obtain the normalized TAM, we apply $relu$ to discard negative values and normalize it by 
\begin{equation}\label{eq:normTAM}
 \textbf{x}_{norm}(I,t) = \frac{\sqrt{relu(\textbf{x}(I,t))}}{\underset{p \in pixels}{max}\sqrt{relu(x(I,t,p))}}.
\end{equation}


\subsection{Direct Estimation of Pixelwise Class Labels from TAMs}\label{sec:inf}
\begin{figure}
 \includegraphics[width=80mm]{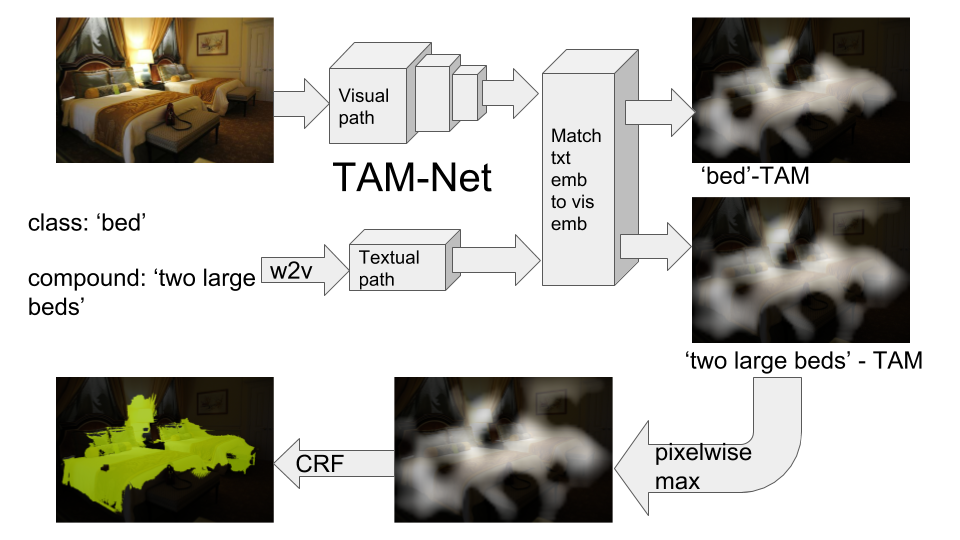}
 \caption{For each present class, we locate the class name as well as all compound concepts related to this class in the image (estimate their TAMs). We then normalize these TAMs and take for each class the pixelwise maximum over them to arive at the CAM. Finally, we estimate pixelwise class labels from these.}
 \label{fig:inf}
  \vspace{-2mm}
\end{figure}

Since we aim to learn a model for semantic segmentation, we need to estimate the pixelwise class labels for each training image $I$. To obtain these, we first calculate normalized class activation maps for all present classes as shown in Figure~\ref{fig:inf}.
To this end, for each class $c$ mentioned in the captions of image $I$, we collect a set of text snippets $\Phi(c)$ which comprise the class name and
all compound concepts related to it. Then we combine the normalized TAMs for all $t\in\Phi(c)$ into a normalized CAM $\textbf{y}_{norm}(I,c)\in\mathbb{R}^{P}$ for class $c$ by taking the pixelwise maximum over the TAMs:
\begin{equation}\label{eq:normCAM}
 y_{norm}(I,c,p) = \underset{t \in \Phi(c)}{max}\{x_{norm}(I,t,p)\} 
\end{equation}
We obtain the background activation map $\textbf{b}(I)$ as in \cite{affinityNet} by
\begin{equation}\label{eq:bgcam}
 b(I,p) = (1 - \underset{c \in C(I)}{max}\{y_{norm}(I,c,p)\})^\alpha
\end{equation}
where $C(I)$ are the classes present in image $I$.
We keep $\alpha=4$ which is the value suggested in \cite{affinityNet}.
We then finally refine the normalized CAMs with a CRF \cite{CRF} to estimate pixelwise class labels.



\subsection{Training of Embedding Architecture}
\begin{figure}
\begin{center}
 \includegraphics[width=60mm]{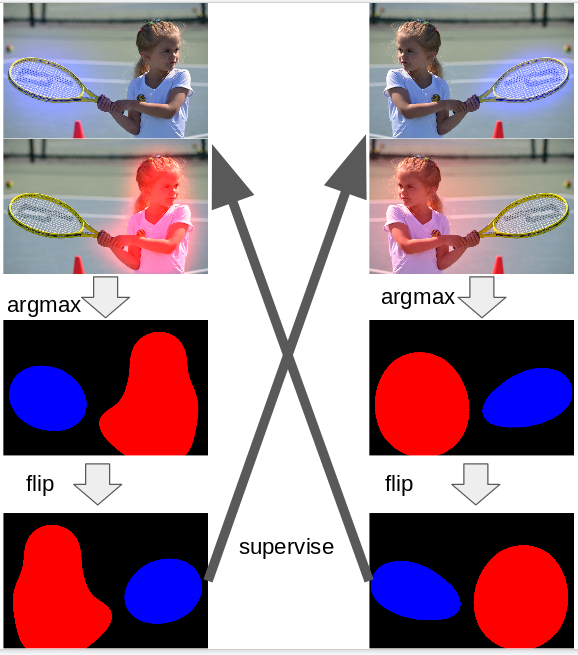}
 \caption{The auto consistency loss enforces invariance under geometric transformations like flipping. To this end, labels are estimated in an online manner from the TAMs of class names for the image and the flipped image. Then these are flipped and the estimated labels of the image are used to supervise the embedding of the flipped image and vice versa.}
 \label{fig:train}
 \end{center}
  \vspace{-5mm}
\end{figure}

We finally describe how we train our network. For training, we propose the following loss  
\begin{equation}\label{eq:loss_sum}
 L = \lambda_{cls}L_{cls} + \lambda_{cpt}L_{cpt} + \lambda_{ac}L_{ac}.
\end{equation}
The class loss $L_{cls}$ and concepts loss $L_{cpt}$ ensure that the textual activation maps have high values for classes and compounds that are present in a training image and low values if they are not present.    
The auto consistency loss $L_{ac}$ ensures that the TAMs are geometric consistent if an image is vertically flipped as illustrated in Figure~\ref{fig:train}.  

\textbf{Class Loss.} 
For each class c, we apply average pooling over the pixels of the TAM for its class name $t_c$, \ie, $x_{pool}(I,t_c)=\frac{1}{P}\sum_p 
\textbf{x}(I,t_c)$. These values can be seen as logits for the probability of this class to be present in the image. We use average pooling as in \cite{paper34, paper4, affinityNet} since it typically leads to activation maps that cover the complete class and not only its most distinctive areas as it is the case for max pooling. 
The class loss $L_{cls}$ is then given by the multilabel binary cross entropy loss:
\begin{equation}
\begin{aligned}
 L_{cls} = & -\sum_{c \in C(I)}log\bigg(\frac{1}{1 + e^{-x_{pool}(I,t_c)}}\bigg) \\
               & -\sum_{c \notin C(I)}log\bigg(\frac{e^{-x_{pool}(I,t_c)}}{1+e^{-x_{pool}(I,t_c)}}\bigg).
 \end{aligned}
\end{equation}
The loss is minimized if the TAMs for the present classes show a strong signal while the TAMs of the classes that are not present in the image are close to zero.       

\textbf{Concepts Loss.}
The concepts loss is calculated in the same way as the class loss, except that we use the TAMs $\textbf{x}(I,t)$ of compound concepts instead of using TAMs of class names. As for the class loss, we apply average pooling $x_{pool}(I,t)=\frac{1}{P}\sum_p 
\textbf{x}(I,t)$. While we use the multilabel binary cross entropy loss as for the class loss, we have to subsample the missing concepts since there are otherwise too many. The concepts loss is then given by   
\begin{equation}
\begin{aligned}\label{eq:conc}
 L_{cpt} = & -\sum_{t \in Comp(I)}log\bigg(\frac{1}{1 + e^{-x_{pool}(I,t)}}\bigg) \\
               & -\sum_{t \in ContrComp(I)}log\bigg(\frac{e^{-x_{pool}(I,t)}}{1+e^{-x_{pool}(I,t)}}\bigg)
 \end{aligned}
\end{equation}
where $Comp(I)$ are the compound concepts present in image $I$ and $ContrComp(I)$ are contrastive compound concepts that are randomly sampled from other images.


\textbf{Auto-Consistency Loss.} The purpose of the auto-consistency loss is to ensure that geometric transformations of the image do not alter the accordingly transformed TAMs as illustrated in Figure~\ref{fig:train}. 
We use the image $I$ and the flipped image $I_{f}$ for training and obtain the corresponding TAMs $\textbf{x}(I,t_c)$ and $\textbf{x}(I_{f},t_c)$ for all classes.
Note that there is no TAM for the background class, we therefore set $\textbf{x}(I,t_{bg})=0$ for the background class.
We convert them into pixel-wise class probabilities $\textbf{Z}$ and $\textbf{Z}_f$ by applying the softmax for each pixel $p$, \ie, $\textbf{Z}(p,c) = \textrm{softmax}_{c'}(\textbf{x}(I,t_{c'},p))$. 

We also compute a pixel-wise labeling for $\textbf{x}(I,t_c)$ and $\textbf{x}(I_{f},t_c)$ as in Section \ref{sec:inf} but without CRF. Instead, we simply use the class with the highest activation per pixel
\begin{equation}
\hat{c}(I, p) = \underset{c \in \{C(I)\cup bg\}}{argmax}x_{norm}(I,t_{c},p)
\end{equation}
where the background activation is estimated as in \eqref{eq:bgcam}. We finally mirror the pixel-wise segmentations as shown in  Figure~\ref{fig:train}. 
The auto-consistency loss $L_{ac}$ is then given by 
\begin{equation}
\begin{aligned}
 L_{ac}(I)= & -\frac{1}{2}\underset{p\in pixel}{\sum}log\bigg(\frac{1}{1+e^{-Z_{f}(p,\hat{c}(I, p))}}\bigg) \\
            & -\frac{1}{2}\underset{p\in pixel}{\sum}log\bigg(\frac{1}{1+e^{-Z(p,\hat{c}_{f}(I, p))}}\bigg).
 \end{aligned}
\end{equation}

\section{Experiments}
For our experiments, we use the COCO dataset \cite{COCO}, since it provides several captions for each image and instance level segmentations for 80 object classes which we convert to class level segmentations.
As train set we use the COCO train2014 split containing 83k images. To evaluate the final semantic segmentation models, we use the COCO val2014 split containing 40k images as our test set.
Our evaluation metric is intersection over union averaged over 81 classes (80 object classes and background).

\textbf{Implementation Details.}
The visual paths of our VGG16 and ResNet38 architecture is identical to the respective architectures from \cite{affinityNet} up to the last layer. While we use mostly the VGG architecture for our experiments, we show some results using the ResNet at the end.   
We train the VGG architecture as well as the ResNet architecture for 15 epochs. For VGG, the learning rate is 0.1 for weights and 0.2 for biases, for ResNet it is 0.01 and 0.02, respectively. For VGG the batch size is 16, for ResNet it is 8. 
Weight decay equals 0.0005 for both architectures. 
The first two convolutional blocks are not finetuned at all and for the fc8 layer and the textual path, we scale up the learning rate by 10. During training, the learning rate decays to 0 according to the polynomial policy. 
The data augmentation techniques include random scaling, cropping and mirroring. 
We set $\lambda_{cls}=1.0$, $\lambda_{cpt}=0.3$ and $\lambda_{ac}=0.001$ so that each loss term is roughly in the same order of magnitude. 

For the concepts loss \eqref{eq:conc}, we sample a maximum of 10 compound concepts mentioned in the captions of an image. 
To collect contrastive compound concepts which are absent in the image, we first sample 10 random images and extract the compound concepts from their captions. Then we randomly sample 50 of these concepts.

For semantic segmentation we use the baseline VGG16 deeplab model \cite{deeplab}.
We keep the hyperparameters but increase the number of iterations by a factor of 3 to account for the bigger size of COCO as compared to Pascal VOC2012.

\subsection{Evaluation of System Components}

\begin{figure*}
\centering
A \textbf{dog} laying on a \textbf{surf board} and riding a small wave \\
  \subfigure{\includegraphics[width=27mm]{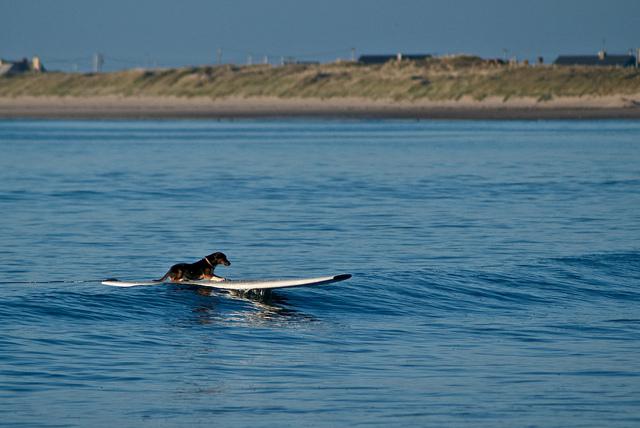}}~~
  \subfigure{\includegraphics[width=27mm]{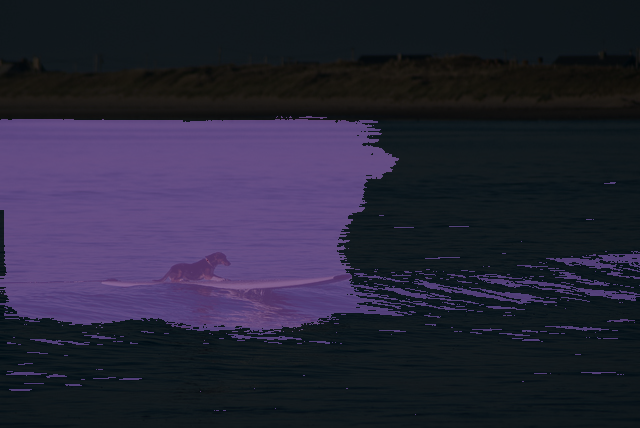}}~~
  \subfigure{\includegraphics[width=27mm]{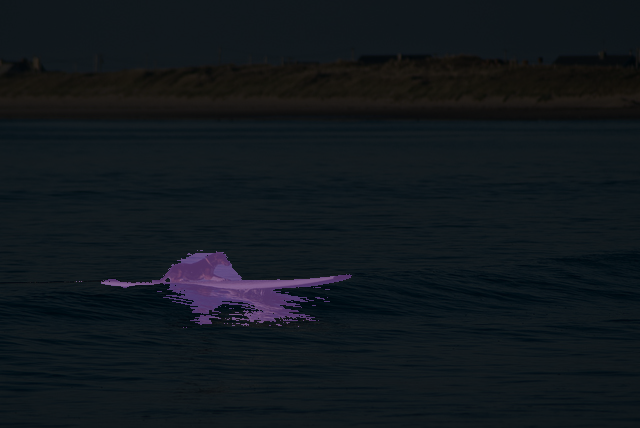}}~~
   \subfigure{\includegraphics[width=27mm]{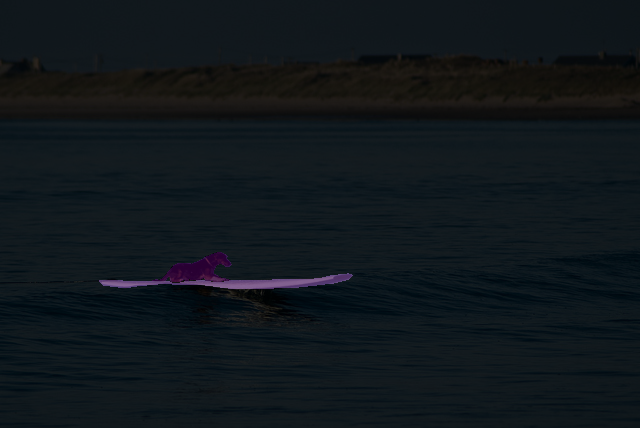}}~~\\
A \textbf{cat} laying on top of a \textbf{bed} next to a window.\\
  \subfigure{\includegraphics[width=27mm]{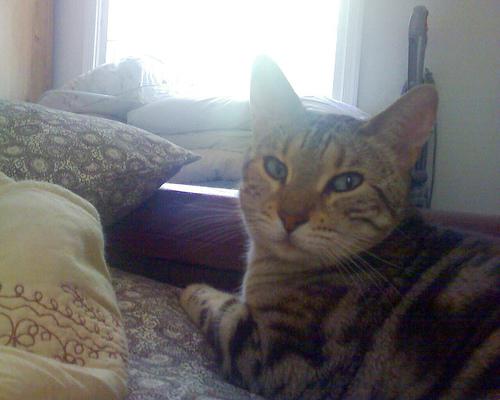}}~~
  \subfigure{\includegraphics[width=27mm]{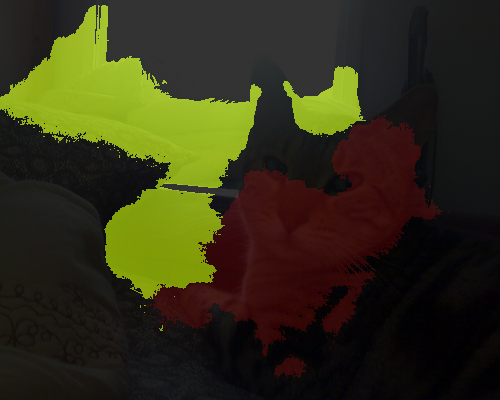}}~~
  \subfigure{\includegraphics[width=27mm]{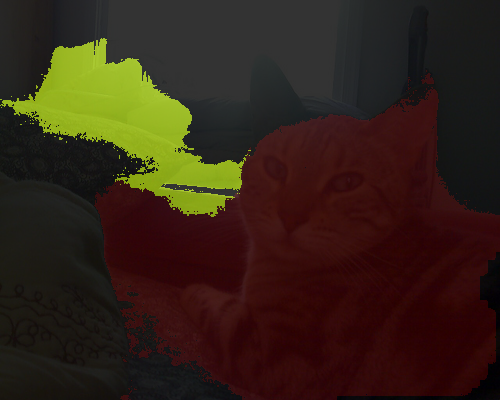}}~~
   \subfigure{\includegraphics[width=27mm]{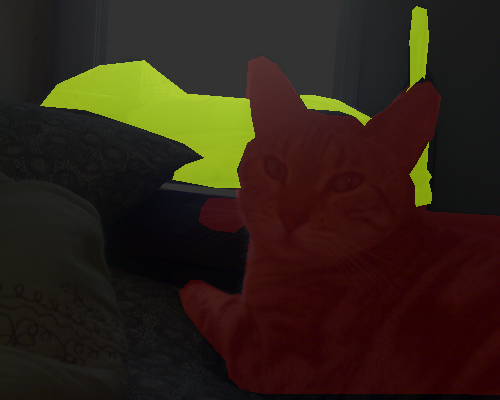}}~~\\
A high-tech \textbf{parking meter} in front of a parked \textbf{car}.\\
  \subfigure{\includegraphics[width=27mm]{COCO_train2014_000000262851.jpg}}~~
  \subfigure{\includegraphics[width=27mm]{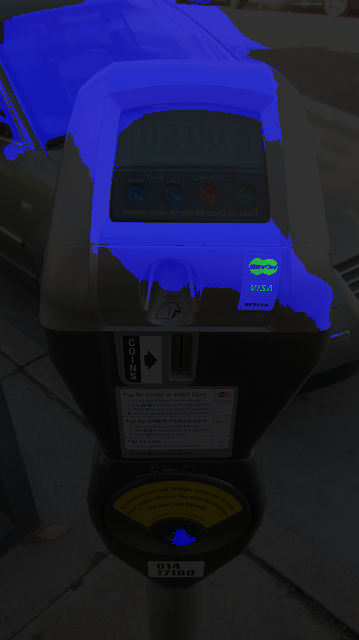}}~~
  \subfigure{\includegraphics[width=27mm]{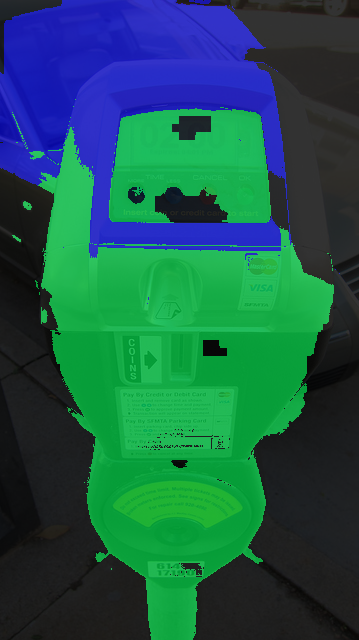}}~~
   \subfigure{\includegraphics[width=27mm]{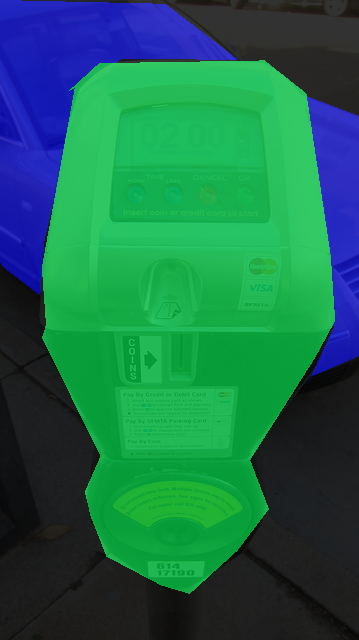}}~~\\
A computer desk, with a \textbf{laptop} on a stand, a \textbf{keyboard}, and a \textbf{mouse}.\\
  \subfigure{\includegraphics[width=27mm]{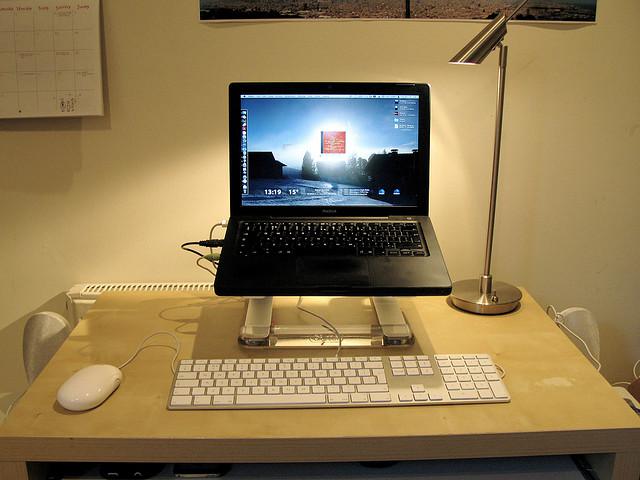}}~~
  \subfigure{\includegraphics[width=27mm]{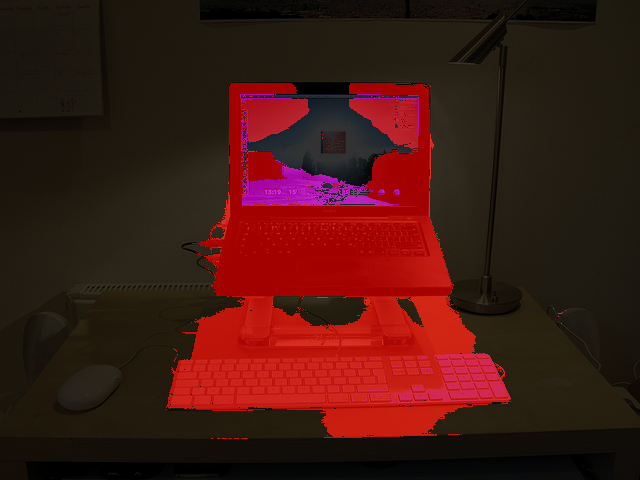}}~~
  \subfigure{\includegraphics[width=27mm]{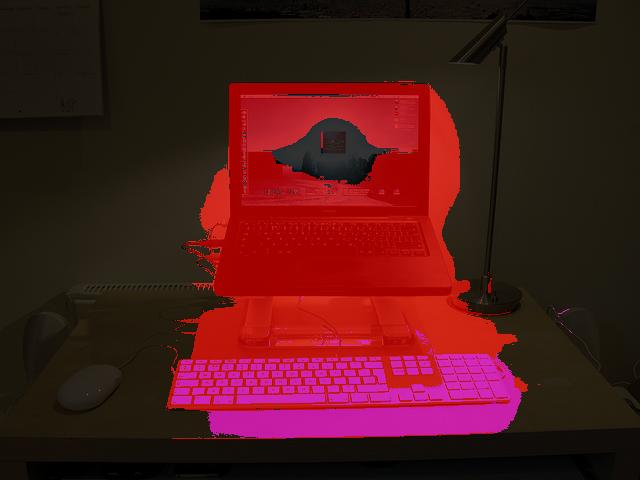}}~~
   \subfigure{\includegraphics[width=27mm]{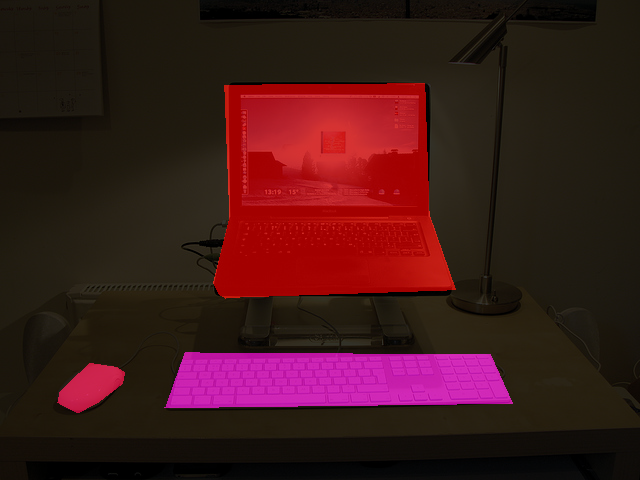}}~~\\
 \caption{Examples of estimated pixelwise class labels on the training set. From left to right: Image, baseline, proposed method, ground truth.
  Above each image we provide the caption and highlight the class names of the COCO classes.}
 \label{fig:dataset_examples}
 \vspace{-2mm}
\end{figure*}

\begin{table*}
\centering
\begin{tabular}{|l|l|l|l|l|l|l|l|}
\hline
\multicolumn{8}{|c|}{Results on training set} \\ \hline
method & TAM-Net & $L_{cls}$ & $L_{cpt}$ & $L_{ac}$ & precision & recall & IoU \\ \hline
\hline
baseline class tags from captions& VGG16 & & & & 0.370 & 0.207 & 0.146 \\ \hline
baseline ground-truth class tags & VGG16 & & & & 0.337 & 0.230 & 0.158 \\ \hline
prop. $L_{cls}$ & VGG16 & $\surd$ & & & 0.378 & 0.205 & 0.144 \\ \hline
prop. $L_{cpt}$  & VGG16& &$\surd$ & & 0.288 & 0.383 & 0.185 \\ \hline
prop. $L_{cls}$ + $L_{cpt}$ & VGG16 &$\surd$ &$\surd$ & & 0.382 & 0.316 & 0.199 \\ \hline
prop. $L_{cls}$ + $L_{cpt}$ + $L_{ac}$  & VGG16 &$\surd$ &$\surd$ &$\surd$ & 0.383 & 0.329 & 0.203 \\ \hline
prop. $L_{cls}$ + $L_{cpt}$ + $L_{ac}$  & ResNet38 &$\surd$ &$\surd$ &$\surd$ & 0.468 & 0.509 & 0.305 \\ \hline
VisGround\cite{DBLP:journals/corr/abs-1804-01720} & ResNet151 & & & & 0.375 & 0.432 & 0.231 \\ \hline
\end{tabular}

\caption{Results for estimated pixel labels on the training set. 
}
\label{tab:CAMs}
\vspace{-2mm}
\end{table*}

For our ablation studies, we first evaluate the accuracy of the pixel labels that are estimated on the training images from the captions   as described in Section~\ref{sec:inf}. 
Simply using a one-hot encoding of the classes retrieved from captions instead of the textual path gives an IoU of 14.6\% as can be seen in Table~\ref{tab:CAMs}. If we use only the class loss $L_{cls}$, the accuracy is similar to the baseline. Using the concepts loss $L_{cpt}$, however, improves mean IoU substantially to 18.5\%. While the recall increases, the precision decreases. This is expected since the compound concepts provide mainly the textual context and are less class focused. Using both loss functions, alleviates this effect and improves the mean IoU to 19.9\%. 
Including auto consistency during training leads to further improvement from 19.9\% to 20.3\%.
We show some qualitative results in Figure~\ref{fig:dataset_examples}.
If we use a ResNet38 architecture instead of a VGG16 architecture, our results improve to 30.5\% IoU.

\subsection{Comparison to Visual Grounding}
We also compare our approach for generating CAMs with the state of the art approach for weakly supervised visual grounding~\cite{DBLP:journals/corr/abs-1804-01720}. 
The authors use complete image captions from the COCO dataset as supervision for training.
During inference, this model receives an image and a single text snippet referring to an object or to stuff in this image and returns an activation map for this snippet.
To segment the image region the snippet refers to, the authors suggest to threshold the activation map with the average of the minimum and maximum value of this map.
We can not use this method out of the shelf since locating multiple text snippets simultaneously is not intended for visual grounding and there is no policy to select the label of a pixel if it is assigned to multiple segments.
To adapt the model to our setting and estimate pixelwise class labels for the training images, we first generate the activation maps for the classes retrieved from the captions. Then, we subtract from each activation map the average of its minimum and maximum value. Finally, we set the activations for background to 0 and apply argmax over the classes. 
This yields 23.1\% IoU as shown in Table~\ref{tab:CAMs}. This is better than our results for the VGG16 architecture but far inferior to our ResNet38 architecture, although the visual grounding model uses a deeper ResNet151 architecture.

\subsection{Evaluating Concept Types}
\begin{table*}
\centering
 \begin{tabular}{|l|l|l|l|l|l|l|l|}
  \hline
\multicolumn{8}{|c|}{Impact of different types of concepts} \\ \hline
type of concepts & cls. rel. comp. cpt. & other comp. cpt. & no adj. & sgl. adj+nouns & precision & recall & IoU \\ \hline
\hline
cls. rel. only &$\surd$ & & & &0.404  &0.306  &0.199  \\ \hline
all comp. conc. &$\surd$ &$\surd$ & & & 0.382 & 0.316 & 0.199 \\ \hline
all concepts &$\surd$ &$\surd$ & &$\surd$ & 0.383 & 0.306 & 0.193 \\ \hline
no adjectives &$\surd$ &$\surd$ &$\surd$ & & 0.380 & 0.283& 0.183 \\ \hline
 \end{tabular}
 \caption{Using compound concepts only for training leads to best results. Results are reported without auto-consistency loss.}
 \label{tab:concept_ablations}
 \vspace{-2mm}
\end{table*}

We compare the performance of our system when using different types of sentence snippets during training and report the results in Table~\ref{tab:concept_ablations}.
In our proposed method, all compound concepts are fed into the concept loss.
Conceptually, adjectives provide additional cues during class activation map calculation. If we discard the adjectives, the accuracy decreases as shown in Table~\ref{tab:concept_ablations}.
Using only compounds containing COCO class names increases the precision but reduces the recall, the IoU remains unchanged.
We therefore use all compound concepts from the captions in all other experiments.
Feeding all nouns and adjectives additionally to the compounds into the loss decreases the performance. 
This shows that the compound concepts better encapsulate the context of the captions than single words.

\subsection{Evaluating Textual Embedding}
\begin{table}
\centering
 \begin{tabular}{|l|l|l|l|l|l|l|}
  \hline
\multicolumn{4}{|c|}{Performance dependence on $w_{res}$} \\ \hline
size of $w_{res}$  & precision & recall & IoU \\ \hline
\hline
0.0 &0.395  &0.287  &0.190  \\ \hline
0.2 (prop.) & 0.382 & 0.316 & 0.199 \\ \hline
0.4  & 0.277 & 0.494 & 0.201 \\ \hline
0.6 & 0.255 & 0.523 & 0.194 \\ \hline
 \end{tabular}
 \caption{Small values of $w_{res}$ allow the textual path to adjust the w2v embeddings to the needs of visual grounding, while large values lead to degeneration during training and inferior performance. Results are reported without auto-consistency loss.}
\label{tab:text_emb}
\vspace{-2mm}
 \end{table}

The hyperparameter $w_{res}$ \eqref{eq:txt} controls the extent to which w2v embeddings \cite{w2v} are adjusted in the textual path. Intuitively high values add flexibility to the network and allow the textual path to adjust the w2v embeddings to the task at hand. Table~\ref{tab:text_emb} reports the results without the auto-consistency loss. As can be seen, increasing $w_{res}$ from 0 improves the performance with a peak at 0.4. However, for higher values, the performance decreases again. This is because higher flexibility allows the network to find degenerate solutions:
If $w_{res}$ becomes too high, the textual path maps all class names and compounds appearing frequently to one vector $v$ and all other class names and compounds to the negative vector $-v$. The visual path then maps everything to $v$.

\subsection{Evaluating Concept Loss Weight}
\begin{table}
\centering
 \begin{tabular}{|l|l|l|l|l|l|l|}
  \hline
\multicolumn{4}{|c|}{Performance dependence on $\lambda_{cpt}$ } \\ \hline
size of $\lambda_{cpt}$  & precision & recall & IoU \\ \hline
\hline
0.1 &0.362  &0.301  &0.186  \\ \hline
0.3 (prop.) & 0.382 & 0.316 & 0.199 \\ \hline
0.5  & 0.396 & 0.319 & 0.204 \\ \hline
 \end{tabular}
 \caption{Performance grows when the impact of compound concepts is increased. Results are reported without auto-consistency loss.}
 \label{tab:conc_weight}
 \vspace{-2mm}
\end{table}
In Table~\ref{tab:conc_weight}, we also evaluate the impact of the parameter $\lambda_{cpt}$ which weights the concept loss in~\eqref{eq:loss_sum}. As in the previous experiment, we do not use the auto consistency loss.
Even a small concept loss already improves the baseline approach giving 18.6\%. By further increasing the weight of the concepts to 0.5, the performances raises to 20.4\%.

\subsection{Comparison to Ground-Truth Image Tags}
The class tags we get from the captions are not perfect as can be seen from Table~\ref{tab:captions}. 
Although for some classes the recall is very low, using parsed tags instead of clean tags does not harm the weakly supervised segmentation performance significantly. Training the baseline model with clean tags instead of retrieved tags improves the mean IoU from 14.6\% to 15.8\% as shown in Table~\ref{tab:CAMs}.
Surprisingly, the precision of CAMs obtained with clean tags is even smaller than with retrieved tags. It seems that COCO object classes not mentioned in the captions are more difficult to locate since captions only mention the most important objects which tend to be large. If an approach for weakly supervised image segmentation fails to locate an object that is expected to be present in an image, the precision decreases.
It is remarkable that the gain from captions is substantially higher than the gain from ground-truth class tags, which shows that the captions provide more information than just tags.
\begin{table}
\centering
\begin{tabular}{|l|l|l|}
\hline
\multicolumn{3}{|c|}{Class tag retrieval precision and recall} \\ \hline
class type & precision & recall  \\ \hline
\hline
person and accessory & 95.0\% & 33.0\% \\ \hline
vehicles & 89.5\% & 63.6\% \\ \hline
outdoor & 95.4\% & 56.0\% \\ \hline
animal & 87.1\% & 92.3\% \\ \hline
sport & 96.2\% & 64.2\% \\ \hline
kitchenware & 89.4\% & 20.3\% \\ \hline
food & 85.9\% & 68.7\% \\ \hline
furniture & 90.3\% & 44.0\% \\ \hline
electronics & 92.3\% & 48.3\% \\ \hline
appliance & 79.8\% & 46.0\% \\ \hline
indoor & 97.0\% & 43.7\% \\ \hline
\end{tabular}
\caption{Recall and precision of image class tags retrieved from image captions.}
\label{tab:captions}
\vspace{-2mm}
\end{table}

\subsection{Results on Test Set}

\begin{table}
\centering
\begin{tabular}{|l|l|l|}
\hline
\multicolumn{3}{|c|}{Results on the test set.} \\ \hline
method & TAM-Net &IoU test set \\ \hline
\hline
Baseline gt class tags no CRF & VGG16 & 0.161 \\ \hline
Proposed no CRF & VGG16 & 0.210\\ \hline
Proposed & VGG16 & 0.216 \\ \hline
Proposed & ResNet38 & 0.285 \\ \hline
\end{tabular}
\caption{Results of the final semantic segmentation model on the test set. The gain in accuracy of estimated pixel labels on the train set transfers well to the test set.}
\label{tab:test_set}
\vspace{-2mm}
\end{table}

To demonstrate the effect of improved CAMs on the final segmentation results, we train deeplab \cite{deeplab} for semantic segmentation on the estimated pixel-wise labels and evaluate its performance on the test set. We first evaluate the impact of the CRF and compare the results obtained by the baseline approach with ground truth class tags to our proposed approach using VGG16 and ResNet38.
The results reported in Table~\ref{tab:test_set} indicate that the quality of the estimated labels is a strong predictor for the quality of the final segmentation. The performance gap of 4.5\% on the train set between the baseline and the proposed method results in 4.9\% on the test set.
Applying a CRF on top of the test set prediction gives a slight improvement of 0.6\%. We use a CRF in all remaining experiments. 
If we use a ResNet38 architecture for our TAM network, we obtain 28.5\%.  

We also compare our method to deep seeded region growing (DSRG)~\cite{paper34}, which is a state of the art weakly supervised semantic segmentation method and it achieves 26.0\% IoU on COCO.
To train their model, the authors take CAMs and background cues from a strongly supervised saliency model as input. We can therefore combine this approach with our method by feeding our generated CAMs to DSRG. We report the results in Table~\ref{tab:DSRG}. For VGG16, DSRG improves IoU from 21.6\% to 26.9\%.
This is also a better result than the number reported by the authors demonstrating the benefit of our CAMs. For ResNet38, however, DSRG decreases the accuracy of our approach. A possible explanation is that DSRG reduces the information from CAMs by estimating high confidence regions first and expands them using the saliency maps. If the CAMs are inaccurate, DSRG substantially improves the results. However, if the CAMs become more accurate, DSRG discards too much information from the CAMs.      

We finally compare our approach with other approaches for weakly supervised semantic segmentation. As can be seen from Table~\ref{tab:SOA}, our approach outperforms the other weakly supervised semantic segmentation models as reported by \cite{paper34}. 
We also include our adapted version of the visual grounding approach of Engilberge et al.~\cite{DBLP:journals/corr/abs-1804-01720}. Interestingly, this approach also achieves a higher IoU than \cite{paper34}, which shows the rich information that is present in image captions. Nevertheless, our approach achieves the highest IoU despite of using a ResNet38 instead of a deeper ResNet151.

\begin{table}
\centering
\begin{tabular}{|l|l|}
\hline
\multicolumn{2}{|c|}{Comparison to DSRG\cite{paper34} } \\ \hline
method &IoU test set \\ \hline
\hline
DSRG\cite{paper34}  & 0.260 \\ \hline
Proposed (VGG16) & 0.216 \\ \hline
Proposed (VGG16) + DSRG\cite{paper34} & 0.269 \\ \hline
Proposed (ResNet38) & {\bf 0.285} \\ \hline
Proposed (ResNet38)+DSRG\cite{paper34} & 0.277 \\ \hline
\end{tabular}
\caption{DSRG\cite{paper34} is a saliency based approach for weakly supervised semantic segmentation. It can be combined with our approach.   
}
\label{tab:DSRG}
\end{table}

\begin{table}
\centering
\begin{tabular}{|l|l|}
\hline
\multicolumn{2}{|c|}{Comparison to the state of the art} \\ \hline
method  &IoU test set \\ \hline
\hline
BFBP\cite{Saleh2016Built} & 0.204 \\ \hline
SEC\cite{SEC} & 0.224 \\ \hline
DSRG\cite{paper34} & 0.260 \\ \hline
VisGround\cite{DBLP:journals/corr/abs-1804-01720} adapted & 0.275 \\ \hline
Proposed & {\bf 0.285} \\ \hline
\end{tabular}
\caption{Comparison of the final semantic segmentation model with the state of the art on the test set.}
\label{tab:SOA}
\vspace{-2mm}
\end{table}
\section{Conclusion}
We presented an approach that uses image captions as supervision for weakly supervised semantic image segmentation. 
Inspired by weakly supervised approaches that estimate class activation maps from class tags and deduce localization cues from them, our approach estimates text activation maps for the class names as well as compound concepts and fuses them to obtain better class activation maps. 
We evaluated our approach on the COCO dataset and demonstrated that our approach outperforms the state of the art for weakly supervised image segmentation.

{\small
\bibliographystyle{ieee}
\bibliography{egbib}
}

\end{document}